# Learning the Dimensionality of Hidden Variables

**Gal Elidan**   **Nir Friedman**
School of Computer Science & Engineering, Hebrew University
{*galel, nir*}@*cs.huji.ac.il*

## Abstract

A serious problem in learning probabilistic models is the presence of *hidden* variables. These variables are not observed, yet interact with several of the observed variables. Detecting hidden variables poses two problems: determining the relations to other variables in the model and determining the number of states of the hidden variable. In this paper, we address the latter problem in the context of Bayesian networks. We describe an approach that utilizes a score-based agglomerative state-clustering. As we show, this approach allows us to efficiently evaluate models with a range of cardinalities for the hidden variable. We show how to extend this procedure to deal with multiple interacting hidden variables. We demonstrate the effectiveness of this approach by evaluating it on synthetic and real-life data. We show that our approach learns models with hidden variables that generalize better and have better structure than previous approaches.

## 1 Introduction

In the last decade there has been a great deal of research focused on the problem of learning Bayesian networks from data (e.g., [11]). An important issue is the existence of *hidden* (*latent*) variables that are never observed, yet interact with observed variables. Hidden variables often play an important role in improving the quality of the learned model and in understanding the nature of interactions in the domain. A crucial problem is the question of how to determine the dimensionality of a hidden variable. This issue is relevant both when learning with fixed structure (e.g., one assessed by an expert) and in cases where the learning algorithm attempts to introduce new variables.

The number of states a hidden variable has can have significant effect on the performance of the model and also on its complexity. For example, Figure 1 demonstrates a common phenomenon: When states of a parent variable $X$ are merged, $X$'s children may no longer be conditionally independent given $X$. As a consequence, more complicated networks, where there are edges among children, might be needed to describe the domain. This phenomenon is more pronounced when the variable $X$ also has parents. The

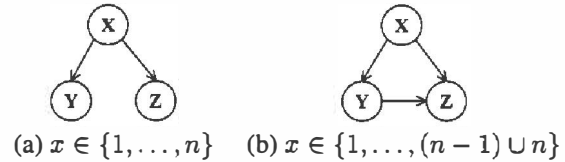

(a) $x \in \{1, \ldots, n\}$    (b) $x \in \{1, \ldots, (n-1) \cup n\}$

Figure 1: Illustration of the change in a network that might results from the merging of two states of a parent variable.

child variables are no longer separated from their ancestors by $X$, and so additional edges are needed. We can see that the correct determination of the cardinality of a hidden variable can affect the complexity of the learned network, which in turn has important ramifications on robustness of learned parameters, and complexity of inference.

In this paper, we propose an agglomerative, score-based approach for determining the cardinality of hidden variables. Our approach starts with the "maximal" number of states possible and merges states in a greedy fashion. At each iteration of the algorithm, it maintains for each training instance a "hard" assignment to the hidden variable. Thus, we can score the data using *complete data* scoring functions that are orders of magnitude more efficient than standard EM-based scores for incomplete data. The procedure progresses by choosing the two states whose merger will lead to the best improvement (or least decrease) in the score. These steps are repeated until all the states are merged into one state. Based on the scores of intermediate stages, we choose the cardinality of the hidden variable. We show that networks learned from the intermediate stages are also good initial starting points for EM runs that fine-tune the parameters.

We then move on to consider networks with multiple hidden variables. As we show, we can combine multiple invocations of the single-variable procedure to learn the interactions between several hidden variable. Finally, we combine our method with the structural detection of hidden variables of Elidan *et al.* [7] and show that this leads to learning better performing models, on test and real-life data.

## 2 Background

### 2.1 Learning Bayesian Networks

Consider a finite set $\mathcal{X} = \{X_1, \ldots, X_n\}$ of discrete random variables where each variable $X_i$ may take states from a finite set, denoted by $Val(X_i)$. A *Bayesian network* is an annotated directed acyclic graph that encodes a joint probability distribution over $\mathcal{X}$. The nodes of the graph correspond to the random variables $X_1, \ldots, X_n$. Each node is annotated with a *conditional probability distribution* (CPD) that represents $P(X_i \mid \mathbf{Pa}_{X_i})$, where $\mathbf{Pa}_{X_i}$ denotes the parents of $X_i$ in $G$. A Bayesian network $B$ specifies a unique joint probability distribution over $\mathcal{X}$ given by:

$$P(X_1, \ldots, X_n) = \prod_{i=1}^{n} P(X_i \mid \mathbf{Pa}_{X_i})$$

The graph $G$ represents conditional independence properties of the distribution. These are the *Markov Independencies*: Each variable $X_i$ is independent of its non-descendants, given its parents in $G$. One implication of the Markov independencies is that a variable $X_i$ interacts directly only with its *Markov Blanket*. This blanket includes the $X_i$'s parents, children, and spouses (additional parents of children of $X_i$). We denote by $\mathbf{MB}_{X_i}$ the variables in the Markov Blanket of $X_i$.

We are interested in learning Bayesian networks from examples. Assume we are given a *training set* $D = \{\mathbf{x}[1], \ldots, \mathbf{x}[M]\}$ of instances of $\mathcal{X}$, that were sampled from an unknown distribution. We want to find a network $B$ that *best matches* $D$. If the structure of the network is given to us, we can use the maximum likelihood approach to estimate the parameters. A more challenging problem is to learn the structure of the network. The common approach to this problem is to introduce a scoring function that evaluates candidate networks with respect to the training data, and then to search for the best network according to this score. A commonly used scoring function to learn Bayesian networks is the *Bayesian scoring* (BDe) metric [12] which we denote by $\text{Score}_{\text{BDe}}$. This scoring metric uses a balance between the likelihood gain of the learned model and the complexity of the network structure representation.

An important characteristic of the score function we use is that when the data is *complete* (that is, each training instance assigns values to all the variables) the score is *decomposable*. More precisely, the score can be rewritten as the sum

$$\text{Score}(G : D) = \sum_i \text{FamScore}_{X_i}(\mathbf{Pa}_{X_i} : D).$$

where the contribution of each variable $X_i$ to the total network score depends only on the states of $X_i$ and $\mathbf{Pa}_{X_i}$ in the training instances. Assuming $\mathbf{Pa}_{X_i} = U$,

$$\text{FamScore}_{X_i}(\mathbf{Pa}_{X_i} : D) = \log P(\mathbf{Pa}_{X_i} = \mathbf{U}) +$$
$$\sum_{\mathbf{u}} \left( \log \frac{\Gamma(\alpha_{\mathbf{u}})}{\Gamma(N[\mathbf{u}] + \alpha_{\mathbf{u}})} + \sum_{x_i} \log \frac{\Gamma(N[x_i, \mathbf{u}] + \alpha_{x_i, \mathbf{u}})}{\Gamma(\alpha_{x_i, \mathbf{u}})} \right)$$

The terms $\alpha_{\mathbf{u}}$ and $\alpha_{x_i, \mathbf{u}}$ are hyper-parameters of the prior distributions over the parameterizations. The terms $N[\mathbf{u}]$ and $N[x_i, \mathbf{u}]$ are *counts* of the number of occurrences of each event in the data. The vector of the counts $N[X_i, \mathbf{Pa}_{X_i}]$ is called a *sufficient statistic* vector for the family $P(X_i \mid \mathbf{Pa}_{X_i})$.

Once we specify the scoring function, the structure learning task reduces to a problem of searching over the combinatorial space of structures for the structure that maximizes the score. The standard approach is to use a local search procedure, such as greedy hill-climbing, that changes one edge at a time.

The learning problem is different when the training data is *incomplete*, that is, some of the states in the training data are missing, or when we learn a network that contains hidden variables that are not observed. In this situation the task is both computationally and conceptually much harder. In order to learn parameters for a given network structure, we can use the *Expectation Maximization (EM)* algorithm to search for a (local) maximum likelihood (or maximum a posteriori) parameter assignment [5, 14].

In the presence of incomplete data, scoring candidate structures is more complex. We cannot efficiently evaluate the marginal likelihood and need to resort to approximations. A commonly used approximation is the *Cheeseman-Stutz* (CS) score [3, 4], which combines the likelihoods of the parameters found by EM, with an estimate of the penalty term associated with structure. The *structural EM* algorithm of Friedman [8] extends the idea of EM to the realm of structure search. Roughly speaking, the algorithm uses an E-step as part of the structure search. The current *model* — structure as well as parameters — is used for computing expected sufficient statistics for other candidate structures. The candidate structures are then scored based on these expected sufficient statistics. The search algorithm moves to a new candidate structure. We can then apply EM again for the new structure, to get the desired expected sufficient statistics and score new candidate structures. This algorithm converges to a "local" maximum. The search space of this algorithm contains many such convergence points, and so care should be taken in choosing the initialization point.

### 2.2 Detecting hidden variables in Bayesian networks

As mentioned in the introduction, we are interested both in cases where the hidden variable is given but its dimensionality is unknown and in constructing new hidden variables. For this purpose, we will use the method for detecting hidden variables that was suggested by Elidan *et al.* [7]. We now briefly review this method.

The general idea of the method is to detect hidden variables by finding *structural signatures* in a Bayesian network learned over the observed variables. As Elidan *et al.* show, the "signature" formed by removing a hidden variable $H$ is a clique over the children of $H$. However, when reconstructing the network from data, we might miss some edges. Thus, instead of searching for perfect cliques, the FindHidden algorithm searches for approximate cliques (relaxation on the number of neighbors) called *semi-cliques*. A semi-clique is a set of variables such that each variable has an edge to at least half of the variables in the *set*.

Once a semi-clique **S** is found, a new hidden variable is proposed. To evaluate this variable, the algorithm constructs a network, with a new variable $H_\mathbf{S}$. This variable is made a parent of the variables in **S**. In addition, all edges among these variables are removed. Then, the algorithm applies a constrained version of structural EM to adapt the structure with $H_\mathbf{S}$ and to estimate parameters for the new network. The score of the learned network is then compared to the score of the original one. The change in score reflects the utility of introducing the hidden variables.

The results of Elidan *et al.* show that this algorithm is successful in introducing hidden variables and improves performance on test data.

## 3 Choosing the Cardinality of a Hidden Variable

We now address the following problem. We are given training data $D$ of samples from $\mathbf{X} = \{X_1, \ldots, X_n\}$, and a network structure $G$ over $\mathbf{X}$ and an additional variable $H$. We need to determine what cardinality of $H$ leads to the best scoring network.

A straightforward way to solve this problem is as follows: We can examine all possible cardinalities of $H$ up to a certain point. For each cardinality $k$, we can apply the EM algorithm to learn parameters for the network containing $H$ with $k$ states. Since EM might get stuck in local maxima, we should perform several EM runs from different random starting points. Given the parameters for the network, we can approximate the score of the network with $k$ states for $H$ using, say, the Cheeseman-Stutz approximation [3]. At the end of the process, we return the cardinality $k$ that received the best score.

This approach is in common use in probabilistic clustering algorithms, e.g., [3]. The central problem of this approach is its exhaustiveness. The EM algorithm is time consuming as it requires inference in the Bayesian network. For simple Naive-Bayes networks that are used in clustering, this cost is not prohibitive. However, in other network structures the cost of multiple EM runs can be high. Thus, we strive to find a method that finds the best scoring cardinality (or a good approximation of it) significantly faster.

We now suggest an approach that works with hard assignments to the states of the hidden variables. This approach is motivated by *agglomerative clustering methods* (e.g., [6]) and *Bayesian model merging* techniques from the HMM literature [17].

The general outline of the approach is as follows. At each iteration we maintain a hard assignment to $H$ in the training data. We can represent this assignment as a mapping $\sigma_H$ from $1, \ldots, M$, to the set $Val(H)$. The assignment, $\sigma_H(m)$ is the state that $H$ holds in the $m'th$ instance. We initialize the algorithm with a variable $H$ that has many states (we describe the details below). We then evaluate the score of the network with respect to the dataset that is

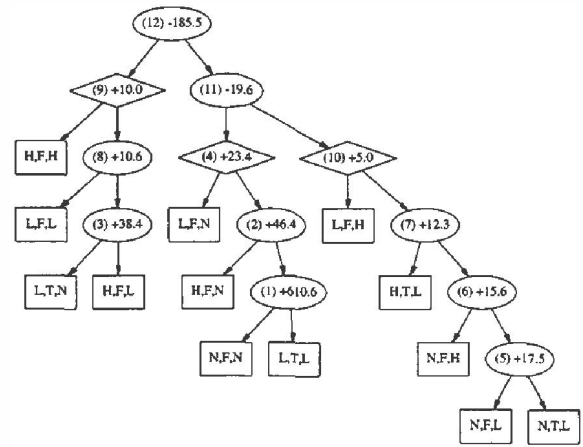

Figure 2: Trace of the agglomeration process in a simple synthetic example. We sampled 1000 instances from the Alarm network, and then hid the observations of the variable *HYPOVOLEMIA* in the data. We then attempted to reconstruct its cardinality. Each leaf in the tree is annotated with the values of the variables in the Markov Blanket (LVEDVOLUME,LVFAILURE and STROKEVOLUME). Nodes correspond to states that result from merging operations. They are numbered according to the order of the merging operations and are annotated with the change in score incurred by the merge operation. Note that at each stage, the merge chosen is the one that produces the largest increase (or smallest decrease) to the score. Diamond-shaped nodes correspond to the final cardinality chosen.

*completed* by $\sigma$. Next, we *merge* two states of $H$ to form a variable with smaller cardinality. This leads to a new assignment function. We then reevaluate the network with respect to this new assignment, and so on. These steps are repeated until $H$ has a single state. We return the number of states $k$ that received the highest score. Figure 2 shows a concrete example of the tree built during such an agglomeration process. We now consider in more the detail the steps in the process.

We start with the initialization point of the algorithm, that is setting the initial states for the variable $H$. Recall that the Markov blanket $\mathbf{MB_H}$ of $H$ separates it from all other variables. This implies that two instances in which $\mathbf{MB_H}$ has the same state, are identical from $H$'s perspective. Thus, the largest number of states that are relevant for a given data sets, is the number of distinct assignments to $\mathbf{MB_H}$ in the data. We initialize $H$ to have a state for each such assignment. In the example of Figure 2 only 13 assignments (out of 16 possible) were observed in the data. We then augment our training data with these assignments to $H$. That is, for each assignment $\mathbf{u} \in Val(\mathbf{MB_H})$, we have a state $h_\mathbf{u}$ and for each instance $m$ we set $\sigma_H(m)$ to be the state $h_\mathbf{u}$ consistent with the Markov blanket assignment of instance $m$.

Once we set $\sigma_H()$, we need to evaluate its usefulness. Since, $\sigma_H()$ assigns a specific state of $H$ for each instance, it *completes* the training data $\sigma_H(D)$. Thus, we can apply a standard complete data score function (e.g., BDe) to our now completed data set. Recall that when the data is complete, $Score_{BDe}$ can be evaluated efficiently as a closed form formula. Moreover, the score depends only on the *sufficient statistics* vectors. Each such vector counts the number of occurrences of each assignment to a variable and its parents. We denote by $N[X_i, \mathbf{Pa}_{X_i}]$ the sufficient statistics that correspond to the family (the node and its parents) of $X_i$ which we denote by $\mathbf{Family}_{X_i}$. To evaluate $\sigma_H$, we only need to consider families that contain $H$: $\mathbf{Family}_H$ and $\mathbf{Family}_C$ for each child $C$ of $H$.

At each iteration of the algorithm we choose to merge two states of $H$, such that the resulting set of states has the best $Score_{BDe}$. Now, suppose that $h_i$ and $h_j$ are two states of $H$ that we want to merge. This means that for all instances where $H$ is assigned $h_i$ or $h_j$, we now assign $H$ to a new state, say $h_{i \cdot j}$. Formally, we define a new function $\sigma'_H$, so that $\sigma'_H(m) = h_{i \cdot j}$ if $\sigma_H(m) = h_i$ or if $\sigma_H(m) = h_j$, otherwise, $\sigma'_H(m) = \sigma_H(m)$. We can then evaluate $\sigma'_H$ and compare its score to the score of $\sigma_H$. This difference is the improvement (or loss) of the merge operation.

We note that when merging states we actually do not need to modify the training data. Instead, we simply apply the merging operation on the sufficient statistics that correspond to $H$ and its children. That is, we set $N[h_{i \cdot j}, \mathbf{pa_H}] = N[\mathbf{h_i}, \mathbf{pa_H}] + N[\mathbf{h_j}, \mathbf{pa_H}]$ for each assignment $\mathbf{pa_H}$ to the parents of $H$. Similarly we compute the sufficient statistics for $H$'s children and their families.

To determine the best merge operation, the algorithm considers all pairs of states of $H$. This can potentially lead to cubic running time (since each iteration require quadratic amount of computation). However, with suitable choice of prior, we can show that the BDe score (and the MDL score, as well) are *locally decomposable*. To make this more precise, suppose that $\sigma'_H$ is the result of merging the states $h_i$ and $h_j$ in $\sigma_H$. Define $\Delta_{i,j} = Score(G : H \cup \sigma'_H) - Score(G : H \cup \sigma_H)$. The score is locally decomposable if $\Delta_{i,j}$ does not depend on other states of $H$. Thus, once we compute this change in score as a result of merging $i$ and $j$, we do not need to recompute it in successive iterations.

A closer look at the properties of the score reveals the behavior we can expect to see when applying our procedure. Recall that the scoring function trades-off between the likelihood of the data and the complexity of the model. When we consider plots of score vs. $H$'s cardinality, we observe three effects that come into play.

1. When merging states of $H$, the number of parameters in the network is reduced. This gives a positive contribution to the score since the complexity of the model is reduced.

2. When $H$ has fewer states, the probability of $H$'s state given its parents is larger. Thus, the likelihood term

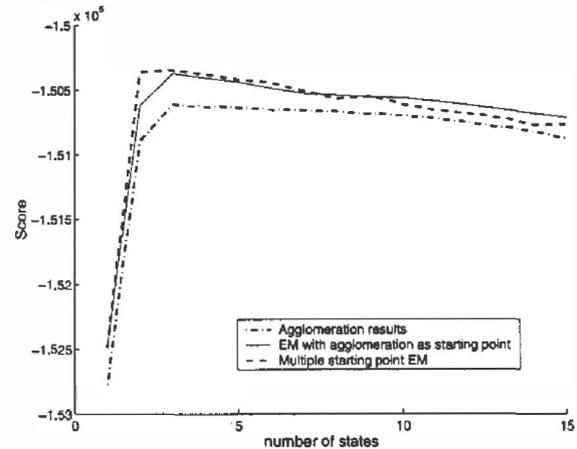

Figure 3: Typical behavior of the score as a function of the number of states in an agglomeration run. BDe score of the agglomeration method, CS score based on an EM run that starts at agglomeration output, and CS score based on the best EM run from multiple starting points. These results shown are for recovering the *STROKEVOLUME* variable in the Alarm network.

associated with $Family_H$ improves after each merge operation.

3. When $H$ has many states, it can provide better prediction of its children. In fact, in our initialization point, $H$'s children are a deterministically determined by $H$'s state (since $H$ has a state for each joint assignment to the Markov Blanket). When the number of states is reduced, the predictions of $H$'s children become more stochastic and their likelihood is reduced. Thus, after a merge, the likelihood of $H$'s children will decrease.

This suggests that the score will increase rapidly due to the contribution of the first two effects, will then slow down but still increase due to the steady contribution of the first effect, and finally decrease and, as we approach a single state, indeed "plunge" due to the third effect.

Figure 3 shows an example of the graph we get when we track the score during iterations of the algorithm. This figure also shows the relations between the score our algorithm assigns to each cardinality $k$ and the one assigned by the standard traditional method that runs EM at each cardinality. In Section 5 we analyze in more detail the two methods.

## 4 Deciding the Cardinality of Several Hidden Variables

In the previous section we examined the problem of deciding the cardinality of a single hidden variable. What happens if our network contains several hidden variables? We start by noting that in some cases, we can decouple the problem: If a hidden variable $H$ is *d-separated* from all the other hidden variables by the observed variables, then we can learn it independent of the rest. More precisely, if $\mathbf{MB_H}$ consists of observable variables only, we do not need to worry about $H$'s interactions with other hidden variables.

However, when two or more hidden variables interact with each other the problem is more complex. A decision about the cardinality of one hidden variable can have effect on the decisions about other hidden variables. Thus, we need to consider a joint decision for all the interacting variables. The standard EM approach mentioned at the beginning of the last section becomes more problematic here since the cardinality space grows exponentially with the number of hidden variables. We now describe a simple heuristic approach that attempts to approximate the cardinality assignment for multiple variables. The ideas are motivated by a similar approach to multi-variable discretization [9].

The basic idea is to apply the agglomerative procedure of the previous section in a round-robin fashion. At each iteration, we fix the number of states and the state assignment to instances for all the hidden variables but one. We apply the agglomerative algorithm with respect to this hidden variable. At the next iteration, we select another variable and repeat the procedure. It is easy to check that we should reexamine a hidden variable only after one of the variables in its Markov Blanked has changed. Thus, we continue the procedure until no hidden variable has changed its cardinality and state assignment.

One crucial issue is the initialization of this procedure. We suggest to start in a network were all hidden variables have one state. Thus, in the initial rounds of the procedure, each hidden variable will be "trained" with respect to its observable neighbors. Only in later iterations, the interactions between hidden variables will start to play a role.

It is easy to see that each iteration of this procedure will improve the score of the "completed" data set specified by the state assignment functions of the hidden variables. It immediately follows that it must converge.

## 5 Experimental Results and Evaluation

We set out to evaluate the applicability of our approach in various learning tasks. We start by evaluating how well our algorithm determines variable cardinality in synthetic datasets where we know the cardinality of the variable we hid. We sampled instances from the Alarm network [1], and manually hid a variable from the dataset. We then gave our algorithm the original network and evaluated its ability to reconstruct the variable's cardinality. Figure 3 shows a typical behavior of the $Score_{BDe}$ vs. the number of states. We repeated this procedure with 24 variables in the Alarm network. (We did not consider variables that were either leafs or had few neighbors.) Using training sets with 10,000 instances, the predictions of cardinality can be broken down as follows:

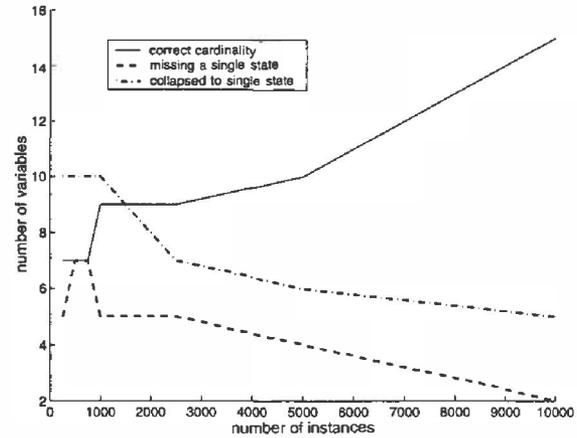

Figure 4: Deviations of the predicted cardinality of the agglomeration method from the true cardinality for 24 variables in the Alarm network as a function of the number of instances. Shown are curves for true cardinality, collapse into a single states and a single missing state (other deviations were rare).

- For 15 variables, the agglomerative procedure recovered the correct cardinality.
- For 2 variables, the estimated cardinality had one state less than the true cardinality.
- For 2 variables, the estimated cardinality had one additional state. Examining the network CPDs suggest that children of these two variables are stochastic in some states of the parents (with almost uniform probability). Initial steps in the agglomeration attempted to model this distribution, which lead to sub-optimal aggregate states in later phases of the agglomeration.
- For 5 variables, the agglomerative procedure suggested a complete collapse into a single state. This is equivalent to removing the variable. A close look at the probabilities in the network shows that these variables have little effect if any on their children and thus they indeed seem almost redundant. In order to confirm this claim, for each of the five variables and for each cardinality, we ran EM from multiple starting points to find the best scoring network. For all the variables, the best score was achieved when the variable was collapsed to a single state.

To summarize, for 19 of 24 of the variables we got the correct or near-perfect prediction of cardinality. For the other 5 variables, the characteristics of the data are two weak to reach statistically significant results.

Next, we tested the effect of the training set size on these decisions. We applied the agglomeration method for all the above variable on training sets with different sizes. Figure 4 shows the deviation from the true cardinality as a function of the training set size. We see that even for small sample

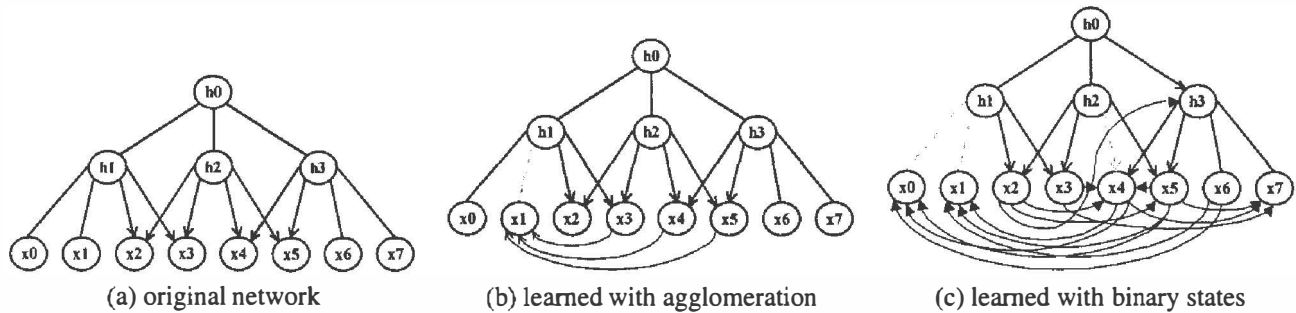

(a) original network　　　　(b) learned with agglomeration　　　　(c) learned with binary states

Figure 5: Performance of the agglomeration algorithm on a network with several interacting hidden variables. Comparison of the model learned with agglomeration (b) to the model learned with binary values (c) demonstrates the importance of determining the cardinality of hidden variables. (dashed light edges are edges that were removed, thin edges are edges that were added)

sizes, the predictions for most variables are either perfect or underestimates the cardinality by 1. This can be expected since the training set does not manifest rare assignments to the Markov blanket of each variable and less states are needed to explain the data.

We then compared our approach to the standard method of evaluating different cardinalities using EM. We compared two variants of EM. The first, performed multiple EM runs from 5 different random starting points. The second variant performed a single EM run starting from the parameters we learn from the "completed" data during the agglomeration step. Figure 3 compares the scores assigned to different cardinalities by the agglomerative approach and these two EM variants for one variable. Note that for all methods the case $k = 3$, which is indeed the correct cardinality, received the highest score. Also note that the two EM variants give similar scores. This suggests that the agglomerative approach finds useful starting points for EM.

In terms of running time, each EM run for each cardinality in this example takes over 250 seconds. The agglomeration procedure takes a little over one second to agglomerate the 15 initial states. One might claim that for determining cardinality, it suffices to run only few iterations of EM, which are computationally cheaper. To test this, we run EM with an early stopping rule. This reduced down the running time of EM about 60 seconds for each run. However, this also resulted in worse estimates of the cardinality, which were worse than these made by the agglomerative method.

We conclude that significant time can be saved by using our method to set the number of states and then apply EM for fine-tuning. This typical behavior was observed in similar comparisons when we hid other variables in the Alarm network.

Next we wanted to evaluate the performance of our algorithm when dealing with multiple hidden variables. To do so, we constructed a synthetic network, shown in Figure 5(a)), with several hidden variables and generated a matching data set with the appropriate variables hidden. Using the true structure as a starting point, we applied our agglomerative algorithm followed by structural EM. As a strawman we also apply a structural EM with binary values for all hidden variables. Because of the flexibility of Structural EM and the challenging structure of our network, we can expect that a learning algorithm that is not precise, will quickly deviate from the true structure. The results are summarized in Figure 5 where $h0, h1, h2$ and $h3$ have 3, 2, 4, and 3 states, respectively, and the visible nodes are all binary. It is evident that the agglomeration method was able to effectively handle several interacting hidden variable. The cardinality was close to the original cardinality with extra states introduces to better explain stochastic relations that do not look stochastic in the training data. The structure learned using the binary model emphasizes the importance of determining the cardinality of hidden variables as suggested in the example of Figure 1. In terms of log-loss score on test data, the model learned with agglomeration was superior to the original model that was better then the model learned with binary values.

We now turn to the incorporation of the cardinality determining algorithm into the hidden variable discovery algorithm of Elidan *et al.* [7] (see Section 2). Given a candidate network, FindHidden searches for semi-cliques and offers candidate hidden variables. It then applies our method to the candidate network to determine the cardinality of the hidden variable. Finally, we allow Structural EM to fine-tune the candidate network.

We applied this to several variables in the synthetic Alarm network. We also experimented on the following real-life data sets: **Stock Data**: a dataset that traces the daily change of 20 major US technology stocks for several years (1516 trading days). These states were discretized to three categories: "up", "no change", and "down". **TB**: a dataset that records information about 2302 tuberculosis patients in the San Francisco county (courtesy of Dr. Peter Small, Stanford Medical Center). The data set contains demographic information such as gender, age, ethnic group, and medical information such as HIV status, TB infection type, and other test results. **News**: data set that contains

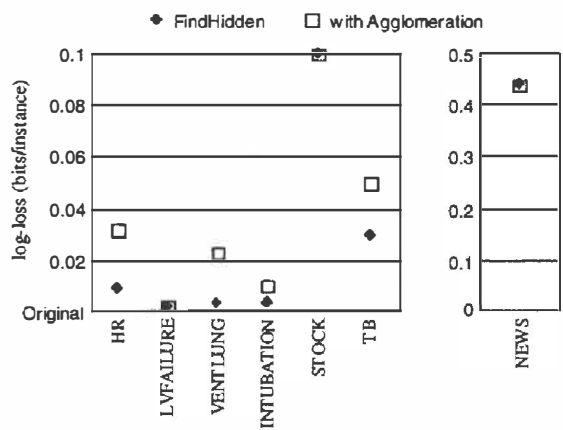

Figure 6: Log-loss performance on test data of the FindHidden algorithm with and without agglomeration on synthetic and real-life data. Base line is the performance of the Original network given as an input to FindHidden

messages from 20 newsgroups [13]. We represent each message as a vector containing one attribute for the newsgroup and attributes for each word in the vocabulary. We removed common stop words, and then sorted words based on their frequency in the whole data set. The data set used here included the group designator and the 99 most common words. We trained on 5,000 messages that were randomly selected from the total data set.

Figure 6 shows the log-loss performance of the networks on test data. The base line is the original network learned without the hidden variable and supplied as input to FindHidden. The solid diamonds are the score of the network with the hidden variable but no agglomeration (hidden variable is arbitrarily set to two states) and the squares are the network with hidden variable with the agglomeration method applied. As we can see, in all cases, the network with the suggested hidden variable outperformed the original network. The network learned using agglomeration performed better then the learned network with no agglomeration (excluding 2 cases where the agglomeration suggested exactly two states and is thus equivalent to the no agglomeration run).

It is interesting to look at the structures found by our procedure. Elidan *et al.* [7] found an interesting model for the TB patient dataset. One state of the hidden variable captures two highly dominant segments of the population: older, HIV-negative, foreign-born Asians, and younger, HIV-positive, US-born blacks. The hidden variable's children distinguished between the two aggregated subpopulations using the *HIV-result* variable, which was also an ancestor of several of them. They noted that it is possible that additional states for the hidden states might have further separated these populations. Figure 7 compares the model learned by the FindHidden algorithm and the model learned with the integration of our agglomerative method. The model does not only perform better on test data (see Figure 6) but does indeed define 4 separate populations: US born, under 30 or over 60, HIV-negative; US born, between 30 and 60 years, with higher probability of HIV; Foreign-born, Hispanics, with some probability of HIV; and Foreign-born, Asians, HIV-negative.

## 6 Discussion and Future Work

In this paper, we proposed an agglomerative, score-based approach for determining the cardinality of hidden variables. We compared our method to the exhaustive approach for setting the cardinality using multiple EM runs and showed its successfulness in generating competing learning models. The importance and plausibility of using the agglomeration method as a pre-processing step to a learning algorithm is an important consequence, thus saving significant computational effort. The algorithm proved robust to the number of instances in the training set. It was also able to deal effectively with several interacting hidden variables. Finally, we evaluated the method as part of the hidden variable detection algorithm FindHidden on synthetic and real-life data and showed improved performance as well as more appealing structures.

Several works are related to our approach. Several authors examined operations of value abstraction and refinement in Bayesian networks [2, 16, 15, 19]. These works were mostly concerned with the ramifications of these operations on inference and decision making. Decisions about cardinality also appear in the context of discretization. Although the data is observable, the introduction of a discretized variable can be modeled as adding a hidden variable. For example, Friedman and Goldszmidt [9] incorporated the discretization process into the learning of Bayesian networks. Like our approach, they use a decomposable score to trade-off between likelihood gain and complexity penalty resulting from a particular discretization. Their approach to discretizing multiple interacting variables is also similar to ours.

In the context of learning hidden variables, the most relevant are the works of Stolcke and Omohundro [17, 18]. In these works, they learn hidden Markov models and probabilistic grammar by performing a bottom up state-agglomeration. Similar to our method, they start by spanning all possible states and then iteratively merging states using information vs. complexity measures. Our work can be viewed as a generalization of their work by applying it to general Bayesian networks and combining it with the hidden variable detection algorithm.

The *structural EM* algorithm of Friedman [8] followed by the work of Elidan *et al.* [7], and with this work are all aimed toward learning non-trivial structures with hidden variables from data. The incorporation of hidden variables is essential both in improving prediction on new examples and to gain understanding of the underlying interactions of the domain. We plan to continue this research project in several directions. We intend to explore additional methods for detecting the dimensionality of hidden variables such as estimating information theoretic measures in situations similar to that of Figure 1. In order to deal effectively with sparse data domains where structural signatures are weak, further methods for the discovery of hidden variable need to be developed. Another direction is to extend the methods for learning hidden structure in more expressive models such as *Probabilistic Relational Models* [10].

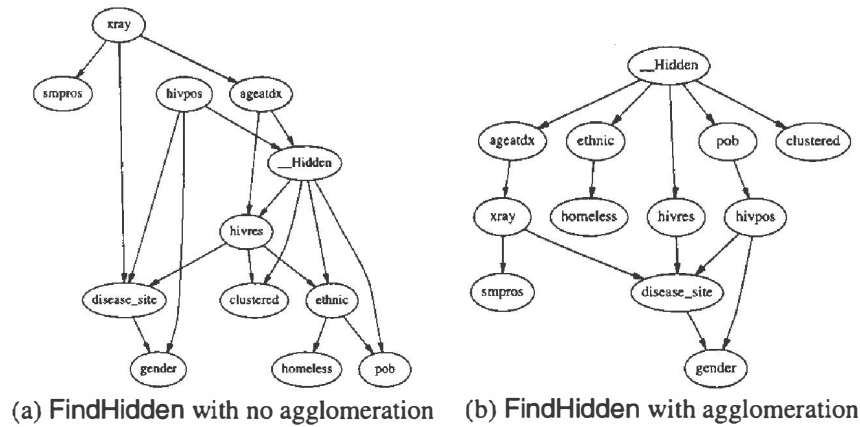

(a) FindHidden with no agglomeration    (b) FindHidden with agglomeration

Figure 7: Improvement in structure of the *TB* network due to incorporation of the cardinality determining algorithm into FindHidden. The hidden variables with 4 states captures more distinct populations and improves the predictive ability of the model.

### Acknowledgements

We thank Noam Lotner and Iftach Nachman for comments on earlier drafts of this paper. This work was supported in part by Israel Science Foundation grant number 224/99-1. Nir Friedman was also supported by an Alon fellowship and the Harry & Abe Sherman Senior Lectureship in Computer Science. Experiments reported here were run on equipment funded by an ISF Basic Equipment Grant.

### References

[1] I. Beinlich, G. Suermondt, R. Chavez, and G. Cooper. The ALARM monitoring system In *Euro. Conf. on AI and Med.*. 1989.

[2] K. Chang and R. Fung. Refinement and coarsening of Bayesian networks. In *UAI '90*, pp. 475–482. 1990.

[3] P. Cheeseman, J. Kelly, M. Self, J. Stutz, W. Taylor, and D. Freeman. AutoClass: a Bayesian classification system. In *ML '88*. 1988.

[4] D. M. Chickering and D. Heckerman. Efficient approximations for the marginal likelihood of Bayesian networks with hidden variables. *Mach. Learning*, 29:181–212, 1997.

[5] A. P. Dempster, N. M. Laird, and D. B. Rubin. Maximum likelihood from incomplete data via the EM algorithm. *J. Royal Stat. Soc.*, B 39:1–39, 1977.

[6] R. O. Duda and P. E. Hart. *Pattern Classification and Scene Analysis*. 1973.

[7] G. Elidan, N. Lotner, N. Friedman, and D. Koller. Discovering hidden variables: A structure-based approach. In *NIPS 13*, 2001.

[8] N. Friedman. The Bayesian structural EM algorithm. In *UAI '98*. 1998.

[9] N. Friedman and M. Goldszmidt. Discretization of continuous attributes while learning Bayesian networks. In *ML '96*, pp. 157–165. 1996.

[10] L. Getoor, N. Friedman, D. Koller, and A. Pfeffer. Learning probabilistic relational models. In *IJCAI '99*. 1999.

[11] D. Heckerman. A tutorial on learning with Bayesian networks. In *Learning in Graphical Models*, 1998.

[12] D. Heckerman, D. Geiger, and D. M. Chickering. Learning Bayesian networks: The combination of knowledge and statistical data. *Mach. Learning*, 20:197–243, 1995.

[13] K. Lang. Learning to filter netnews. In *ML '95*, pp. 331–339, 1995.

[14] S. L. Lauritzen. The EM algorithm for graphical association models with missing data. *Comp. Stat. and Data Ana.*, 19:191–201, 1995.

[15] K. Poh, M. Fehling, and E. Horvitz. Dynamic construction and refinement of utility based categorization models. *IEEE Trans. Sys. Man Cyb.*, 24(11):1653–1663, 1994.

[16] K. Poh and E. J. Horvitz. Reasoning about the value of decision-model refinement: Methods and application. In *UAI '93*, pp. 174–182. 1993.

[17] A. Stolcke and S. Omohundro. Hidden Markov Model induction by Bayesian model merging. In *NIPS 5*, pages 11–18. 1993.

[18] A. Stolcke and S. Omohundro. Inducing probabilistic grammars by Bayesian model merging. In *Inter. Conf. Grammatical Inference*, 1994.

[19] M. P. Wellman and C.-L. Liu. State-space abstraction for anytime evaluation of probabilistic networks. In *UAI '94*, pp. 567 – 574, 1994.